\documentclass[10pt,twocolumn,letterpaper]{article}

\usepackage{cvpr}
\usepackage{times}
\usepackage{epsfig}
\usepackage{amsmath}
\usepackage{amssymb}
\usepackage{graphicx} 
\usepackage{booktabs}
\usepackage[breaklinks=true,bookmarks=false]{hyperref}

\cvprfinalcopy %

\def\cvprPaperID{****} %

\begin{document}

\title{Pythia v0.1: the Winning Entry to the VQA Challenge 2018}

\author{Yu Jiang$^*$, Vivek Natarajan$^*$, Xinlei Chen$^*$, Marcus Rohrbach, Dhruv Batra, Devi Parikh\\
Facebook AI Research
}

\maketitle

\renewcommand*{\thefootnote}{\fnsymbol{footnote}}
\footnotetext{$^*$ indicates equal contributions.}
\renewcommand*{\thefootnote}{\arabic{footnote}}
\newcommand{\myquote}[1]{\emph{`#1'}}

\begin{abstract}
This document describes Pythia v0.1, the winning entry from Facebook AI Research (FAIR)'s A-STAR team to the VQA Challenge 2018\footnote{and changes made after the challenge deadline.}. 

Our starting point is a modular re-implementation of the bottom-up top-down (up-down) model \cite{anderson2017bottom,teney2017tips}. We demonstrate that by making subtle but important changes to the model architecture and the learning rate schedule, fine-tuning image features, and adding data augmentation, we can significantly improve the performance of the up-down model on VQA v2.0 dataset ~\cite{balanced_vqa_v2} -- from 65.67\% to 70.24\%. 

Furthermore, by using a diverse ensemble
of models trained with different features and on different
datasets, we are able to significantly improve over the `standard' way of ensembling (\ie same model with different random
seeds) by 1.31\%. Overall, we achieve 72.27\% on the test-std split of the VQA
v2.0 dataset. Our code in its entirety (training, evaluation, data-augmentation, ensembling) and pre-trained models are publicly available at: \url{https://github.com/facebookresearch/pythia} .
\end{abstract}

\section{Introduction}

\vspace{-20pt}
\begin{align*}
\text{Chaerephon: \quad} & \textit{Pythia -- Is there any man alive} \\[-0.2em] 
 & \textit{wiser than Socrates?} \\[-0.2em]
\text{Pythia: \quad} & \textit{None}. 
\end{align*}

We present Pythia v0.1, a modular framework for Visual Question Answering research, which formed the basis for the winning entry to the VQA Challenge 2018 from Facebook AI Research (FAIR)'s 
A-STAR\footnote{Agents that See, Talk, Act, and Reason.} team.  

The motivation for Pythia comes from the following observation -- a majority of today's Visual Question Answering (VQA) models fit a particular design paradigm, with modules for question encoding, image feature extraction, fusion of the two (typically with attention), and classification over the space of answers. The long-term goal of Pythia is to serve as a platform for easy and modular  
research \& development in VQA~\cite{antol2015vqa} and related directions like visual dialog~\cite{visdial}. The name \myquote{Pythia} is an homage to the Oracle of Apollo at Delphi, who answered questions in Ancient Greece.

The starting point for Pythia v0.1 is a modular reimplementation of the bottom-up top-down (up-down) model \cite{teney2017tips}. In this study, we demonstrate that by making a sequence of subtle but important changes, we can significantly improve the performance as summarized in Table 1\footnote{FAIR A-STAR's entry in the VQA 2018 Challenge was 72.25\%. This document describes results produced by our code release which reaches 72.27\%.}.

\section{Bottom-Up and Top-Down Attention}
We perform ablations and augmentations over the baseline system of the up-down model~\cite{anderson2017bottom}, which was the basis of the winning entry to the 2017 VQA challenge. The key idea in up-down is the use of an object detector -- Faster RCNN~\cite{ren2015faster} pre-trained on the Visual Genome dataset~\cite{krishna2017visual} -- to extract image features with bottom-up attention, \ie, visual
feed-forward attention. Specifically, a ResNet-101 was chosen as the backbone network, and its entire Res-5 block was used as the second-stage region classifier for detection. After training, each region was then represented by the 2048$D$ feature after average pooling from a $7{\times}7$ grid.

The question text is then used to compute the top-down attention, \ie, task specific attention, for each object in the image. Multi-modal fusion is done through a simple Hadamard product followed by a multi-label classifier using a sigmoid activation function to predict the answer scores. Their performance reached 70.34\% on VQA 2.0 test-std split with an ensemble of 30 models trained with different seeds. For presentation clarity, we present our proposed changes (and the respective improvements) in a sequence; however, we also found them to be independently useful.

\subsection{Model Architecture}
\label{sec:model}
We made a few changes to the up-down model to improve training speed and accuracy. Instead of using the gated hyperbolic tangent activation~\cite{anderson2017bottom}, we use weight normalization~\cite{salimans2016weight} followed by ReLU to reduce computation\footnotemark. We also replaced feature concatenation with element-wise multiplication to combine the features from text and visual modalities when computing the top-down attention. To compute the question representation, we used 300$D$ GloVe~\cite{pennington2014glove} vectors to initialize the word embeddings and then passed it to a GRU network and a question attention module to extract attentive text features \cite{yu2018beyond}. For fusing the image and text information, we found the best-performing hidden size to be 5000. With these modifications, we were able to improve the performance of the model from 65.32\% to 66.91\% on VQA v2.0 test-dev. 

\subsection{Learning Schedule} \label{LearningSchedule}
Our model is optimized by Adamax, a variant of Adam with infinite norm~\cite{kingma2014adam}. In one popular implementation of up-down\footnotemark[\value{footnote}] 
\footnotetext{\url{https://github.com/hengyuan-hu/bottom-up-attention-vqa}} 
learning rate is set to 0.002 with a batch size of 512. We found that reducing the batch size improves performance -- which suggests that there is potential for improving performance by increasing the learning rate. However, naively increasing the learning rate resulted in divergence. To increase the learning rate, we thus deployed the warm up strategy~\cite{goyal2017accurate} commonly used for large learning-rate training of networks.
Specifically, we begin with a learning rate of 0.002, linearly increasing it at each iteration till it reaches 0.01 at iteration 1000. 
Next, we first reduce the learning rate by a factor of 0.1 at 5$K$ and then reduce it every 2$K$ iterations, and stop training at 12$K$. With this we increase the performance from 66.91\% to 68.05\% on test-dev.

\begin{table}[t]
    \centering
    \caption{\label{tab:results} Accuracy (\%) on VQA v2.0. 
    For ease of presentation, our changes are presented as a sequence building on top of 
    previous changes. 
    $^*$ denotes that these models are {\it not} included in our ensemble results submitted to the challenge.%
    }
    \vspace{0.1cm}
    \begin{tabular}{lcc}
    \toprule
    Model & test-dev  & test-std  \\
    \cmidrule(lr){1-3}
       up-down \cite{anderson2017bottom}  &65.32 & 65.67\\
       
        up-down Model Adaptation (\S\ref{sec:model}) & 66.91&\\
        + Learning Schedule (\S\ref{LearningSchedule}) & 68.05&\\
        \ \ \ \ + Detectron \& Fine-tuning (\S\ref{sec:finetune}) & 68.49&\\
        \ \ \ \ \ \ \ \ + Data Augmentation$^*$ (\S\ref{sec:dataaug}) & 69.24 &\\ 
        \ \ \ \ \ \ \ \ \ \ \ \ + Grid Feature$^*$ (\S\ref{sec:gridfeat}) & 69.81 &\\
        \ \ \ \ \ \ \ \ \ \ \ \ \ \ \ \ + 100 bboxes$^*$ (\S\ref{sec:gridfeat}) & 70.01 & 70.24
         \\ \cmidrule(lr){1-3}
        Ensemble, 30$\times$ same model (\S\ref{sec:ensembing}) & 70.96&\\
        Ensemble, 30$\times$ diverse model (\S\ref{sec:ensembing}) & 72.18 & 72.27\\
        \bottomrule
    \end{tabular}
\end{table}

\subsection{Fine-Tuning Bottom-Up Features}
\label{sec:finetune}
Fine tuning pre-trained features is a well known technique to better tailor the features to the task at hand and thus improve model performance~\cite{ren2015faster}. 

Different from Anderson \etal \cite{anderson2017bottom}, we also used the new state-of-the-art detectors based on feature pyramid networks (FPN)~\cite{lin2017feature} from Detectron\footnote{\url{https://github.com/facebookresearch/Detectron}}, which uses ResNeXt ~\cite{Xie2016} as backbone and has two fully connected layers (fc6 and fc7) for region classification. This allows us to extract the 2048$D$ fc6 features and fine-tune the fc7 parameters, as opposed to the original up-down~\cite{anderson2017bottom}, where fine-tuning previous layers requires significantly more storage/IO and computation on $7{\times}7{\times}2048$ convolutional feature maps. 
Similar to up-down, we also used Visual Genome (VG)~\cite{krishna2017visual} with both objects and attributes annotations to train the detector. 

We set the fine-tune learning rate as 0.1 times the overall learning rate. 
We are able to reach a performance of 68.49\% on test-dev with this fine-tuning.

\subsection{Data Augmentation}
\label{sec:dataaug}
We added additional training data from Visual Genome~\cite{krishna2017visual} and Visual Dialog (VisDial v0.9)~\cite{visdial} datasets. For VisDial, we converted the 10 turns in a dialog to 10 independent question-answer pairs. Since both VG and VisDial datasets only have a single ground-truth answer while VQA has 10, we simply replicated the answer to each question in VG and VisDial 10 times to make the data format compatible with the VQA evaluation protocol.  

We also performed additional data augmentation by mirroring the images in the VQA dataset. We do some basic processing of the questions and answers for the mirrored images by interchanging the tokens ``left'' and ``right'' in the questions and answers which contain them. When adding these additional datasets, we reduce the learning rate as we described in Section \ref{LearningSchedule} first at 15$K$ iterations, respectively, and stop training at 22$K$ iterations. As a result of data augmentation, we are able to improve our single model performance from 68.49\% to 69.24\% on test-dev.

\begin{figure}[t]
\vspace{-0.2cm}
\includegraphics[width=1\linewidth]
                   {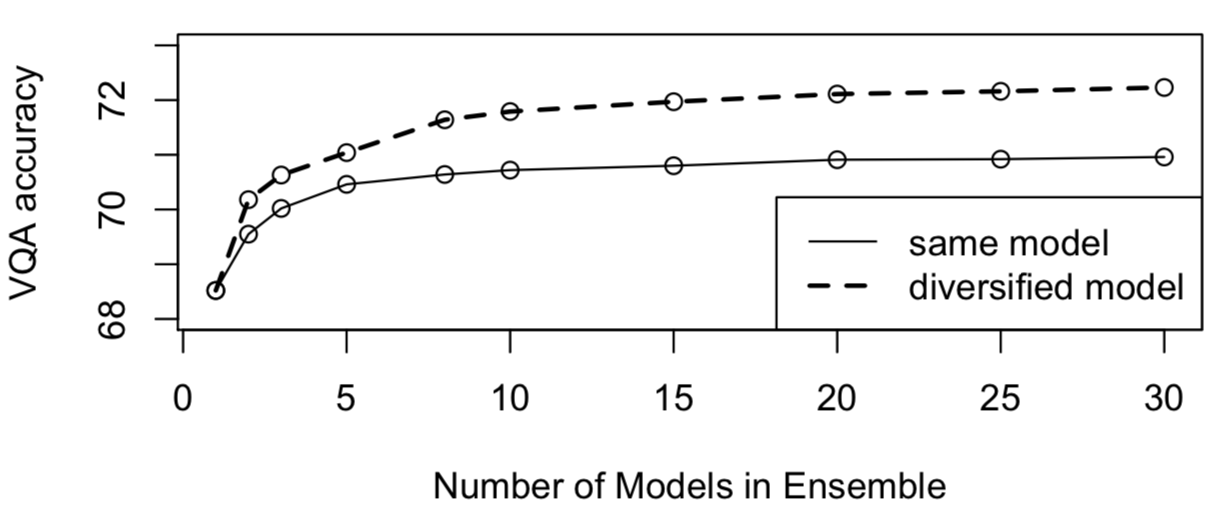} \vspace{-0.5cm}
\caption{\label{fig:ensemble}Performance with different ensemble strategies.}
\end{figure}

\subsection{Post-Challenge Improvements}
\label{sec:gridfeat}
Anderson \etal \cite{anderson2017bottom} uses only 
the features pooled from object proposals (called bottom-up features) to represent 
an image. Our hypothesis is that such a representation does not fully capture 
a holistic spatial information about the image and visual representations 
from image regions not covered by the proposals. 
To test this hypothesis, we combined grid-level image features 
together with bottom-up features. We follow the same procedure as ~\cite{fukui16mcb} to extract grid-level features from ResNet152~\cite{he2016deep}. Object-level features and grid-level features are separately fused with features from questions and then are concatenated to fed to classification. 
Before the challenge deadline, we had experimented with this only on images from the VQA dataset without fine-tuning. After the challenge, we performed more comprehensive experiments and found that adding grid level features helps to further improve the performance to 69.81\%. 

Instead of using an adaptive protocol for choosing the number of object proposals 
(between 10 and 100) per image as as done in \cite{teney2017tips}, we also experimented with using 
a simpler (but slower) strategy of using 100 objects proposals for all images. As can be seen in Table~\ref{tab:results}, with features from 100 bounding-boxes, we reach 70.01\% for test-dev and 70.24\% for test-std on VQA 2.0.

\subsection{Model Ensembling}
\label{sec:ensembing}
All ensembling experiments described below involve models trained {\it before} the challenge deadline. That is, they do not include the two after-challenge experiments described in Section~\ref{sec:gridfeat}. We tried two strategies for ensembling. First, we choose our best single model and train the same network with different seeds, and finally average the predictions from each model.  As can be seen from Fig~\ref{fig:ensemble}, the performance plateaus at 70.96\%. Second, we choose models trained with different settings, \ie, the tweaked up-down model trained on the VQA dataset with/without data augmentation and  models trained with image features extracted from different Detectron models with/without data augmentation. 
As can be seen, this ensembling strategy is much more effective than the previous one. Ensembling 30 diverse models, we reach 72.18\% on test-dev and 72.27\% on test-std of VQA v2.0.

\section*{Acknowledgements}
We would like to thank Peter Anderson, Abhishek Das, Stefan Lee, Jiasen Lu, Jianwei Yang, Licheng Yu, Luowei Zhou for helpful discussions, Peter Anderson for providing training data for the Visual Genome detector, Deshraj Yadav for responses on EvalAI related questions, Stefan Lee for suggesting the name \myquote{Pythia}, Abhishek Das, Abhishek Kadian for feedback on our codebase and Meet Shah for making a docker image for our demo.

{\small
\bibliographystyle{ieee}

}

\end{document}